\pdfoutput=1

\documentclass[11pt]{article}

\usepackage{ACL2023}

\usepackage{times}
\usepackage{latexsym}

\usepackage[T1]{fontenc}

\usepackage[utf8]{inputenc}

\usepackage{microtype}

\usepackage{inconsolata}

\usepackage{enumitem}

\usepackage{graphicx}
\usepackage{caption}
\usepackage{subcaption}
\usepackage{arydshln}

\usepackage{algorithm}
\usepackage{algpseudocode}

\usepackage{amsmath}
\DeclareMathOperator*{\argmax}{\arg\max}

\usepackage{multirow}

\algnewcommand\algorithmicforeach{\textbf{for each}}
\algdef{S}[FOR]{ForEach}[1]{\algorithmicforeach\ #1\ \algorithmicdo}

\title{Minimal Evidence Group Identification for Claim Verification}

\author{Xiangci Li\textsuperscript{\rm 1}$^*$ ~~ Sihao Chen\textsuperscript{\rm 2}$^*$ ~~ Rajvi Kapadia\textsuperscript{\rm 3} ~~ Jessica Ouyang\textsuperscript{\rm 1} ~~ Fan Zhang\textsuperscript{\rm 3}\\
  \textsuperscript{\rm 1} University of Texas at Dallas, 
  \textsuperscript{\rm 2} University of Pennsylvania \\
  \textsuperscript{\rm 3} Google Research\\
  {\tt lixiangci8@gmail.com, sihaoc@cis.upenn.edu, rajvikapadia@google.com} \\
  {\tt Jessica.Ouyang@utdallas.edu, zhanfan@google.com} \\
}

\begin{document}
\maketitle
\def\thefootnote{*}\footnotetext{~Work performed while the authors are interning at Google}\def\thefootnote{\arabic{footnote}}
\begin{abstract}

Claim verification in real-world settings (e.g. against a large collection of candidate evidences retrieved from the web) typically requires identifying and aggregating a complete set of evidence pieces 
that collectively provide full support to the claim.
The problem becomes particularly challenging when there exists distinct sets of evidence that could be used to verify the claim from different perspectives. In this paper, we formally define and study the problem of identifying such \emph{minimal evidence groups} (MEGs) for claim verification. 
We show that MEG identification can be reduced from Set Cover problem, based on entailment inference of whether a given evidence group provides full/partial support to a claim.  Our proposed approach achieves 
18.4\% \& 34.8\% absolute improvements on the WiCE and SciFact datasets over LLM prompting.
Finally, we demonstrate the benefits of MEGs in downstream applications such as claim generation.

\end{abstract}

\vspace{-1em}
\section{Introduction} \label{sec:introduction}

\begin{figure}[t]
\centering
  \includegraphics[width=\linewidth]{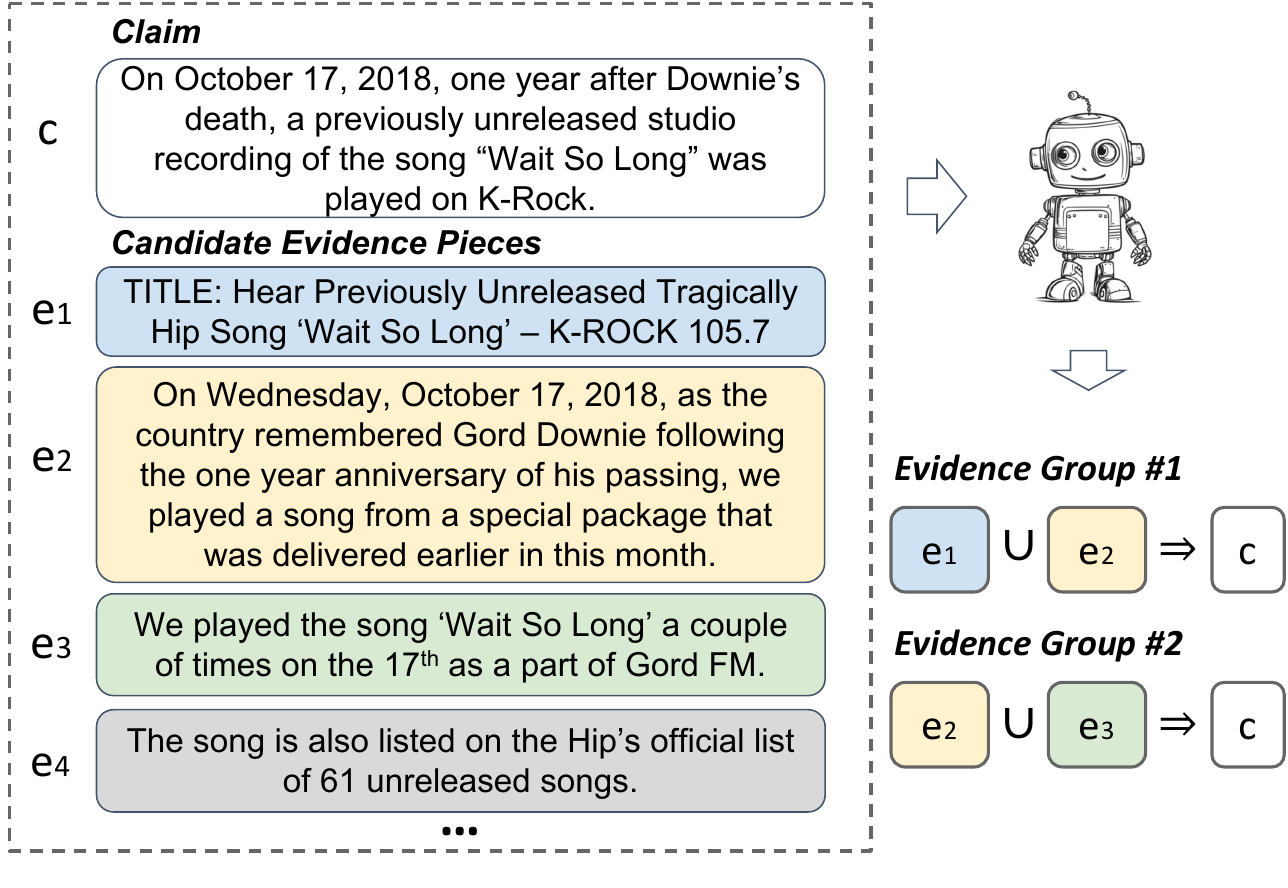}
\caption{The problem of \emph{minimal evidence group} identification for claim verification: given a claim and a list of candidate evidence pieces, the task is to identify the sets of \emph{minimal, non-redundant} evidence, where each set provides full support for the claim.}
  \label{fig:illustration}
  \vspace{-1em}
\end{figure}

The task of \textit{claim verification} predicts whether a claim is supported by the presented evidence \cite{thorne-etal-2018-fever, chen2023complex}. 
A claim verification model is expected to identify the correct evidence pieces (EPs; e.g. evidence sentences or snippets) among tens of retrieved candidate evidence, but a practical challenge lies in that there might exist multiple sets of evidence that verify the claim from different perspectives. Figure~\ref{fig:illustration} shows an example where, given a claim and some retrieved evidence, there exist two different --- but both valid --- ways of supporting the claim. 

While humans can quickly identify mutually redundant EPs, e.g. $e_1$ and $e_3$ in Figure~\ref{fig:illustration}, and propose plausible combinations of EPs as \textit{evidence groups} (EGs, formally defined in Section \ref{sec:formulation}), existing claim verification systems \cite{dagan2005pascal, thorne-etal-2018-fever, wadden-etal-2020-fact, schuster-etal-2021-get, chen2023complex, chen-etal-2023-propsegment} focus only on the relationship between the claim and individual EPs, without considering the co-supporting relationships among EPs. This becomes problematic because the retrieved EPs might be redundant, or an individual EP may only partially support the claim. An EG with redundant EPs makes it more difficult to explain the reasoning for supporting the claim, while an EG composed of partially supporting EPs may still not fully support the claim, resulting in logical flaws. These problematic outputs not only confuse human verifiers, but also hurt the performance of downstream tasks.

In this paper, we introduce the problem of identifying \emph{minimal evidence groups} (MEGs) from retrieved evidence candidates. 
Conceptually, an MEG is composed of EPs with the following properties: (1) \textbf{Sufficiency:} each MEG fully supports the veracity of the claim; (2) \textbf{Non-redundancy:} the EPs in an MEG are not redundant with each other; and (3) \textbf{Minimality:} the number of EPs in each MEG is minimal. 
We formally define the task of MEG identification 
and show that classic claim verification approaches cannot effectively solve this problem. We propose a simple yet practical approach to decompose it to support prediction and evidence group merging. Our proposed approach significantly outperforms the baseline of directly prompting a large-language model (LLM) by 18.4\% and 34.8\% absolute precision scores on the WiCE \cite{kamoi-etal-2023-wice} and SciFact \cite{wadden-etal-2020-fact} benchmarks. Finally, we demonstrate the benefit of MEGs for saving computation budget in the downstream task of claim generation. 


\section{Related Work} \label{sec:related_work}

Classic claim verification \cite{thorne-etal-2018-fever, chen2023complex} consists of three steps: evidence retrieval, evidence selection, and stance prediction. \textit{Evidence retrieval} perform coarse-grained filtering of EPs from thousands of candidates. \textit{Evidence selection} and \textit{stance prediction} perform fine-grained selection of EPs and predict whether the claim is supported by the selected EPs. MEG identification builds on classic claim verification by restricting evidence selection and stance prediction to predict a minimal group of EPs that fully supports the claim. 

To address the problem that claim verification systems \cite{dagan2005pascal,wadden-etal-2020-fact, schuster-etal-2021-get, chen-etal-2023-propsegment} may predict EPs that only partially support the claim, 
\citet{laban-etal-2022-summac, schuster-etal-2022-stretching, kamoi-etal-2023-wice} aggregated individual EPs' entailment scores into EG scores. However, they did not address the problem of redundancy within an EG; we propose MEG identification to fill this gap.

The closest work to ours is SciFact \cite{wadden-etal-2020-fact}, which annotates ``minimal evidence sets'' for each claim. However, the SciFact shared task does not penalize non-minimal EGs, and consequently models that evaluate on SciFact
\cite{pradeep-etal-2021-scientific, li2021paragraph, zhang-etal-2021-abstract, wadden-etal-2022-multivers} are trained on the union of EGs from different human annotators, which is no longer minimal. Similarly, \citet{thorne-etal-2018-fever, chen-etal-2023-propsegment, kamoi-etal-2023-wice} collect (possibly redundant) EGs from multiple annotators and use their union as training labels.
As a result, existing models prioritize EP recall and are not directly comparable to MEG identification models.

\section{Minimal Evidence Groups}
\subsection{Problem Formulation} \label{sec:formulation}
MEG identification builds on the classic claim verification task \cite{thorne-etal-2018-fever, chen2023complex}. Formally, claim verification takes a claim $c$ and a list of candidate EPs $E=\{e_1, e_2, ...\}$ as input. The evidence selection step retrieves all EPs that are \textit{relevant} to $c$, and the stance prediction step predicts whether the selected EPs support $c$\footnote{We limit our scope to claim support/non-support, ignoring contradictions for simplicity. See Section \ref{sec:experiments} for discussion.}. 
In Figure~\ref{fig:illustration}, $ e_1, e_2, e_3$ all support $c$. A set of fully supporting EPs is called an evidence group (EG).

MEG identification requires the EGs to be \textit{sufficient}, \textit{non-redundant}, and \textit{minimal}. We consider a set of EPs $S \subseteq E$ to \textit{fully} or \textit{partially support} a claim $c$ if the EPs in $S$ collectively entail all or only part of $c$, respectively; $S$ does \textit{not support} $c$ if none of EPs in $S$ entail $c$. If $S$ \textit{fully supports} $c$, it is an EG; an MEG is a minimal EG such that none of its EPs are redundant in terms of supporting $c$.
In Figure~\ref{fig:illustration}, $e_1$ and $e_3$ are redundant; $\{e_1, e_2\}$ and $\{e_2, e_3\}$ are MEGs that fully support $c$.

\subsection{Task Evaluation} \label{sec:eval_metrics}
We focus on precision-oriented scores (precision and $F_{0.5}$) to penalize predicting non-minimal EGs because we observe from prior claim verification datasets \cite{thorne-etal-2018-fever, wadden-etal-2020-fact,chen-etal-2023-propsegment, kamoi-etal-2023-wice} that (1) one MEG is sufficient for claim verification in practice; (2) humans are good at finding one plausible MEG but struggle to exhaustively find all valid MEGs; and (3) different annotators propose distinct MEGs. 

Given a claim $c$ with reference MEGs $RG=\{G_1, G_2, ...\}$, we measured the following metrics:

\textbf{Exact match of MEGs} treats each reference MEG atomically and considers a predicted MEG to be correct if it exactly matches a reference MEG. 


\textbf{Best soft match of MEGs} gives partial credits to the predicted MEGs. We calculate the EP-level scores between the predicted MEG $G'$ and the most similar reference MEG chosen by $\hat{G}=\argmax_{G_i\in RG} F_{0.5}(G', G_i)$.


\setlength{\textfloatsep}{0pt}

\begin{algorithm}[t] \small
\caption{MEG identification with a support prediction \textit{Model}. Simplified for illustration, see Appendix Section \ref{sec:implementation} for details.}\label{alg:evidence_group_identification_illustration}
\begin{algorithmic}
\Require $c$, $E=\{e_1, e_2, ..., e_n\}$, $Model$
\Require $max\_size$ \Comment{Max size of EGs to consider.}
\State $MEG \gets []$  \Comment{Proposed MEGs.}
\For{$size$ in $1 ... \min(|E|, max\_size)$}
\State $CS \gets makeCombinations(c, E, size)$ \Comment{List of \textit{notRedundant} combinations of partially supporting EPs.}
\For{$S$ in $CS$}
\State $label \gets Model(c, S)$ 
\If{$label$ is $fully\ support$}
    \State $MEG.append(S)$
     
\EndIf
\EndFor
\If {$len(MEG)>0$}
\textbf{break}
\EndIf
\EndFor
\State \textbf{Output}  $MEG$
\end{algorithmic}
\end{algorithm}

\section{MEG Identification Approach} \label{sec:approach}
The challenge of MEG identification is to find the smallest set of EPs that fully supports the claim. As discussed in Section \ref{sec:related_work}, classic claim verification models treat the EP as the basic unit; they are neither designed nor trained for \textit{groups} of evidence. 
Our experiments of prompting directly with LLMs also show poor performance (Table \ref{tab:llm_minimum_evidence_group}, ``Direct'')
\footnote{The explicit verification of combinations of EPs reduces from Set Cover and is NP-hard (see proof in Appendix \ref{sec:proof}.)}. 

As Algorithms \ref{alg:evidence_group_identification_illustration} shows, 
we decompose MEG identification into two steps: (1) predicting whether a candidate set of EPs \textit{fully supports}, \textit{partially supports}, or does \textit{not support} the claim and (2) bottom-up merging \textit{partially supporting} groups in search of a \textit{fully supporting} group.
The support prediction \textit{Model} can be implemented by any reasonable approach, such as prompting LLMs or fine-tuning models like T5 \cite{2020t5}.
When merging groups, we increment the overall group size by 1 at each step.
Note that if we only evaluate the base case with $size$=1, this is equivalent to classic claim verification \cite{thorne-etal-2018-fever, wadden-etal-2020-fact, schuster-etal-2021-get, kamoi-etal-2023-wice}. 

Based on the definition of MEG (Section \ref{sec:formulation}), we derive three principles to prune the problem space for a tractable solution:
(1) any superset of an MEG \textit{fully supports} the claim $c$;
(2) any non-empty subset of an MEG \textit{partially supports} $c$; 
and (3) if a set of EPs $S$ \textit{fully supports} or does \textit{not support} $c$, then $S$ is not a strict subset of any MEG.
Therefore, we stop merging sets that are predicted as \textit{fully supporting} or \textit{not supporting} to maintain the non-redundancy and minimality of the candidate EP sets. In addition, when choosing a pair of sets to merge, we prune the candidate merge partners for each set using a redundancy checker \textit{notRedundant} (implemented as a zero-shot LLM prompt; see Appendix \ref{sec:implementation}). Finally, upon finding a \textit{fully supporting} set, we stop merging and return all \textit{fully supporting} sets of the current \textit{size}.

\section{Intrinsic Evaluation} \label{sec:experiments}
\subsection{Experimental Settings}

\subsubsection{Datasets}
We perform filtering to convert classic claim verification datasets to align with our MEG identification task. Both of the datasets listed below annotate EGs with multiple annotators. We assume that every human-annotated EG \textit{fully supports} its claim, every subset of an EG \textit{partially supports} its claim, and all non-labeled sentences do \textit{not support} the claim. In addition, we assume each reference EG to be an MEG proposed by a different annotator.

\paragraph{SciFact \cite{wadden-etal-2020-fact}} is a biomedical fact-checking dataset and is the only existing dataset whose annotation instructions match the sufficiency, non-redundancy, and minimality requirements of MEGs. We remove all examples whose claims \textit{contradict} the evidence
, resulting in 268 samples from the development set. We use the non-contradictory EGs as-is. To distinguish it from the original SciFact dataset and task\footnote{As discussed in Section \ref{sec:related_work}, while the SciFact dataset annotates EGs that meet the requirements of MEGs, the task does not evaluate redundancy or minimality, only sufficiency.}, we call this modified dataset \textbf{SciFact-MEG}.

\paragraph{WiCE \cite{kamoi-etal-2023-wice}} distinguishes EGs that \textit{fully} or \textit{partially support} claims from Wikipedia. We use the subclaim-level partition of the dataset and only use samples labeled as \textit{fully supporting}, resulting in 528 samples from the test set. We call this modified dataset \textbf{WiCE-MEG}.

\subsubsection{Implementation} 
For both the support prediction \textit{Model} and \textit{notRedundant} checker, we prompt PaLM-2L \cite{anil2023palm} with few-shot demonstrations and greedy decoding (see Appendix \ref{sec:base_model}). We follow \citet{wan-etal-2023-universal} to select the LLM's most confident examples for few-shot demonstrations. To prioritize precision, we take the top-1 predicted MEG, ranked according to the LLM's predicted \textit{fully supporting} score, if multiple MEGs are predicted.

\subsubsection{Baseline Approaches} 
\textbf{Direct prediction.} We zero-shot prompt PaLM-2L \cite{anil2023palm} to predict the MEG via EP indices, given a claim and a list of candidate EPs (Appendix Table \ref{tab:direct_prompt}). 

\textbf{Classic claim verification.} To simulate classic claim verification without considering groups of EPs \cite{thorne-etal-2018-fever, wadden-etal-2020-fact, schuster-etal-2021-get, kamoi-etal-2023-wice}, we use our proposed approach but early stop after computing $size$=1. If we find any \textit{fully supporting} EP, the output MEG will be the same as our proposed approach. Otherwise, we concatenate all \textit{partially supporting} EPs as a single EG.

\textbf{Classic claim verification with less redundancy (Classic+LR).} Given the output from ``classic claim verification'' above, we additionally remove EPs that cause redundancy, as predicted by the pair-wise \textit{nonRedundant} checker\footnote{We simply remove redundant combinations when $size$=2.}.

\begin{table}[t]
\begin{center}
\setlength{\tabcolsep}{1pt} 
\small
    \begin{tabular}{  c | c | c | c c c }
    \hline
    & & \textit{Exact Match} &  \multicolumn{3}{c}{\textit{Best Soft Match}} \\
    \textbf{Dataset} & \textbf{Approach} & \textbf{Precision} & \textbf{Prec.} & \textbf{Recall} & $\mathbf{F_{0.5}}$ \\ \hline
    \multirow{4}{*}{WiCE-MEG} & Direct & 0.456 & 0.176 & 0.522 & 0.203 \\ \cline{2-6} 
     & Classic & 0.568 & 0.338 & \textbf{0.554} & 0.367 \\ \cline{2-6} 
     & Classic+LR & 0.570 & 0.425 & 0.526 & 0.442 \\ \cline{2-6} 
     & Ours & \textbf{0.640} & \textbf{0.809} & 0.423 & \textbf{0.684} \\ \hline \hline
    \multirow{4}{*}{SciFact-MEG} & Direct & 0.243 & 0.235 & \textbf{0.652} & 0.269\\ \cline{2-6} 
     & Classic & 0.479 & 0.468 & 0.478 & 0.470\\ \cline{2-6} 
     & Classic+LR & 0.479 & 0.491 & 0.476 & 0.488\\ \cline{2-6} 
     & Ours & \textbf{0.591} & \textbf{0.612} & 0.352 & \textbf{0.533}\\ \hline 
    \end{tabular}
    \vspace{-0.5em}
    \caption{Top-1 minimal evidence group identification performance. Examples with failed outputs are excluded for the baseline approach.} \label{tab:llm_minimum_evidence_group}
\end{center}
\end{table}

\subsection{Experimental Results}

\label{sec:evidence_group_performance}
Table \ref{tab:llm_minimum_evidence_group} shows the top-1 MEG identification performance using the metrics introduced in Section \ref{sec:eval_metrics}. 
For both datasets, our approach significantly outperforms all baselines on precision and $F_{0.5}$ scores. The baselines underperform our approach because their predicted MEGs contain too many EPs, especially the ``Direct" LLM prompting baseline. Decomposing MEG identification into many individual entailment problems (``Classic'') greatly improves the precision score. Further removing pair-wise redundancy (``Classic+LR") slightly improves performance, showing the impact of redundancy. Finally, although requiring significantly more computation, our bottom-up MEG identification approach performs the best because every combination of EPs is explicitly verified.


\section{Extrinsic Evaluation}
The non-redundancy of MEGs not only makes the evidence more human-readable, it also improves the performance of downstream applications. Inspired by \citet{chen2023dense}, we use WiCE-MEG to highlight the MEG's minimality and sufficiency properties using claim generation as an example downstream task, with a computation budget measured in the number of words or sentences. 

\subsection{Experimental Settings} \label{sec:budgeted_RAG_settings}
Since EGs fully entail their claims, they contain the information to reconstruct the claim. We compare the following input settings for the task of claim reconstruction using PaLM-2L \cite{anil2023palm}:

\textbf{MEGs.} We use the top-1 MEG obtained with our proposed approaches
, each baseline in Table \ref{tab:llm_minimum_evidence_group}, and the human-annotated reference EG with the smallest number of EPs for each claim. 

\textbf{Union of EGs (UEGs).} We take the union of reference EGs (from different annotators) for a claim. 

\textbf{First-$k$.} To simulate a low computation budget setting, we follow \citet{chen2023dense} in taking the first $k$ EPs, where $k$ is the size of the top-1 MEG.


\begin{table}[t]
\begin{center}
\setlength{\tabcolsep}{3pt} 
\small
    \begin{tabular}{  c | c c | c c c }
    \hline
    \textbf{Input Evidence} & \textbf{\# Words} & \textbf{\# Sents} & \textbf{R-1} & \textbf{R-2} & \textbf{R-L} \\ \hline
    Direct & 172.4 & 6.81 & 0.299 & 0.127 & 0.263 \\ \hline
    First-k Direct & 34.1 & 1.15 & 0.282 & 0.114 & 0.250 \\ \hline 
    Classic & 85.0 & 3.20 & 0.282 & 0.120 & 0.250 \\ \hline
    Classic+LR & 69.2 & 2.45 & 0.281 & 0.120 & 0.250 \\ \hline
    Our MEGs & \textit{39.5} & \textit{1.29} & \textit{0.289} & \textit{0.121} & \textit{0.254} \\ \hline\hline
    Gold MEGs & \textit{37.0} & \textit{1.31} & \textit{0.294} & \textit{0.126} & \textit{0.259} \\ \hline
    Gold UEGs & 71.7 & 2.78 & 0.302 & 0.128 & 0.267 \\ \hline
    First-$k$ gold UEGs & 33.0 & 1.31 & 0.264 & 0.101 & 0.232 \\ \hline
    \end{tabular}
    \vspace{-0.5em}
    \caption{Budgeted retrieval-augmented generation performance (ROUGE F1).} \label{tab:budgeted_RAG}
\end{center}
\end{table}

\subsection{Experimental Results}
Table \ref{tab:budgeted_RAG} shows that both our predicted and gold MEG settings perform comparably to settings with much lower computation budgets, while significantly outperforming the most constrained ``first-$k$'' settings. These results indicate that (1) our proposed approach for MEG identification is effective; (2) MEGs contain complete information for the claim generation task; (3) MEGs are much more compact than EGs from other approaches, with more than 45\% fewer words, allowing them to be used in low-computation scenarios. 




\section{Conclusion}
We have addressed the challenging scenario in claim verification where a model is expected to identify a minimal group of evidence pieces (EPs) among a relatively large amount of candidate evidence, and there might exist different sets of evidence that verify the claim from different perspectives.
We formally define and study the problem of such minimal evidence group (MEG) identification and show that it can be reduced from a Set Cover-like problem. Our proposed approach achieves significant improvements over direct LLM prompting. Finally, we demonstrate the benefit of MEGs over classic claim verification approaches in downstream applications such as claim generation.

\section*{Limitations}
\paragraph{Ignoring contradictions.} In this work, we only consider supporting/non-supporting evidence for simplicity, and leave contradicting evidence for future work. Our proposed approach avoids the edge case of full/partial entailment problem brought by contradiction.
Nonetheless, we claim that contradiction can be regarded as the opposite of support, where our proposed concepts and approaches still apply with minor fix.


\paragraph{Reliability of human annotations.} As we point out in Section \ref{sec:introduction}, there is no gold-standard annotated dataset designed for this task, and it is practically difficult to enforce and verify the sufficiency, non-redundancy, and minimality requirements of MEGs in the existing annotations. In practice, unless explicitly stated, it is unknown whether the annotated EGs are simply relevant to or fully support the claim. Although human annotators are good at proposing salient EGs, annotators usually do not exhaustively find all possible EGs. Moreover, human annotators tend to over-select EPs for a better contextualization, which breaks the non-redundancy and minimality requirements. As a result, we argue that the human annotations should only be treated as a reference, instead of an absolute gold standard. 
Therefore, the measured performance in Table \ref{tab:llm_minimum_evidence_group} can should be regarded as each approach's agreement with the human annotators, and does not necessarily measure MEG correctness. 

\paragraph{Definition of \textit{partial support}.} It is challenging to precisely define \textit{partial support}. Even \citet{kamoi-etal-2023-wice}, who proposed this label, did not clearly define it. Our proposed approaches do not rely on the precise definition of \textit{partial support} but simply regard it as the intermediate label between \textit{not support} and \textit{fully support} because the precise definition may vary case-by-case in different datasets that the support prediction \textit{Model} is trained on. 
Because of this ambiguity, \textit{partial support} is the most challenging label for LLMs to predict (Appendix \ref{sec:base_model}) and hurts the performance of MEG identification.

\paragraph{Running time.} Due to the NP-hardness (Appendix \ref{sec:proof}) of the MEG identification problem, we trade off running time for higher performance, thus the worst case running time for the proposed solution is too long to be practically useful in a production system. Our proposed approach is currently more suitable for dataset creation, which requires a robust solution without strict running time requirements. We leave more efficient approaches for future work.

\section*{Ethical Statements}
Similar to all prior claim verification works \cite{dagan2005pascal, thorne-etal-2018-fever, wadden-etal-2020-fact, schuster-etal-2021-get, chen2023complex, chen-etal-2023-propsegment}, we stress that the MEG identification problem and the MEGs predicted by our proposed approach only consider the relative entailment relationship between the evidence and the claim. In other words, the MEG identification problem and our proposed approach do not guarantee the absolute correctness of the claim or the EPs or EGs themselves. Any future application must be cautious in distinguishing between retrieving evidence that supports the claim, correct or not, and verifying the absolute factual correctness of the claim.

\clearpage
\bibliography{custom}
\bibliographystyle{acl_natbib}
\clearpage
\appendix

\begin{table*}[t]
\begin{center}
\small
    \begin{tabular}{ l | l | l l l | l l l | l l l l }
    \hline
    & \textbf{Accuracy} & \multicolumn{3}{l}{\textbf{Precision}} & \multicolumn{3}{l}{\textbf{Recall}} & \multicolumn{4}{l}{\textbf{F1}}  \\
    \textbf{Dataset} & & \textit{Full} & \textit{Partial} & \textit{Not} & \textit{Full} & \textit{Partial} & \textit{Not} & \textit{Full} & \textit{Partial} & \textit{Not}  & \textit{Macro F1}\\ \hline
    WiCE & 0.792 & 0.891 & 0.373 & 0.960 & 0.790 & 0.612 & 0.866 & 0.838 & 0.464 & 0.911 & 0.737 \\ \hline
    SciFact & 0.729 & 0.833 & 0.077 & 0.794 & 0.741 & 0.095 & 0.848 & 0.784 & 0.085 & 0.820 & 0.563 \\ \hline
    \end{tabular}
    \vspace{-0.5em}
    \caption{Base model performance.} \label{tab:base_model}
    \vspace{-1.5em}
\end{center}
\end{table*}



\section{Proof of NP-hardness}
\label{sec:proof}
In this section, we provide a simple proof to show that the MEG identification problem is NP-hard.

\subsection{Simplifying to an Ideal Scenario} \label{sec:ideal_formulation}
Inspired by \citet{kamoi-etal-2023-wice}, who break complicated claims into subclaims and verify each subclaim individually, we assume the solution of the MEG identification problem explicitly breaks down the claim $c$ into one or more atomic \textit{claim units} $CU=\{cu_1, cu_2, ...\}$ and verifies them one-by-one. Each claim unit $cu$ can be more fine-grained or abstractive than the subclaims introduced by Kamoi et al. If all claim units $cu_i \in CU$ are verified, then $c$ is \textit{fully supported}. Otherwise, if only a subset of $CU$ is verified, then $c$ is only \textit{partially supported}. In an ideal scenario, we have a perfect model that is able to decompose $c$ into $CU$ and output a binary vector for each EP to indicate which $cu_i$ are verified by the EP; this \textit{ideal MEG identification problem} becomes the task of minimizing the number of selected EPs such that all elements in $CU$ can be covered.

\subsection{Reduction from Set Cover}
Based on the formulation in \ref{sec:ideal_formulation}, we can trivially many-one reduce the Set Cover problem, which is NP-Complete \cite{karp2010reducibility}, to \textit{ideal MEG identification} by mapping the universe to $CU$ and the collection of subsets to the full set of EPs $E=\{e_1, e_2, ...\}$. 
Therefore the \textit{ideal MEG identification problem} is NP-Complete, and the actual MEG identification problem is NP-hard. Because the assumption of explicitly tracking which $cu_i$ are covered/remaining is challenging for state-of-the-art models, it is difficult to develop approximation solutions for MEG identification.

\begin{table}[t]
\small
\begin{center}
\setlength{\tabcolsep}{5pt} 
    \begin{tabular}{p{0.9\linewidth} }
    \hline
    \textbf{Prompt} \\ \hline
    Your task is to examine if the given claim is jointly supported by one or more evidence with short contexts.
    Take a deep breath and reason step by step, and answer with ``FULLY\_SUPPORTED'', ``PARTIALLY\_SUPPORTED'' or ``NOT\_SUPPORTED'' at the end of your answer.
    FULLY\_SUPPORTED means the claim is fully supported by the evidence without requiring other evidence.
    PARTIALLY\_SUPPORTED means the claim is partially covered by the evidence that requires other evidence to collectively fully support the claim.
    NOT\_SUPPORTED means the claim is not supported by the evidence.\\
    \\
    Example: \\
    Claim: \{\{example claim\}\} \\
    Evidence with contexts: \\
    \{\{example evidence text\}\} \\
    Answer: \{\{example answer\}\} \\
    \\
    Example: \\
    ...\\
    Your problem: \\
    Claim: \{\{claim\}\} \\
    Evidence with contexts: \\
    \{\{evidence text\}\} \\
    Answer: \\
   \hline
    \end{tabular}
    \vspace{-0.5em}
    \caption{Prompt for \textit{base problem}.} \label{tab:base_problem_prompt}
\end{center}
\end{table}

\begin{table}[t]
\small
\begin{center}
\setlength{\tabcolsep}{5pt} 
    \begin{tabular}{p{0.9\linewidth} }
    \hline
    \textbf{Prompt} \\ \hline
    Each of the following two evidence individually partially support the claim: ``\{\{claim\}\}''.\\
    Partial support means the claim is partially supported by the evidence that requires other evidence to collectively fully support the claim.\\
    \\
    Evidence 1: ``\{\{evidence text 1\}\}''.\\
    Evidence 2: ``\{\{evidence text 2\}\}''.\\
    \\
    Are evidence 1 and 2 redundant to each other in terms of how they support the claim, i.e. are they talking about the same thing, and is one of the evidence unnecessary?\\
    Take a deep breath and think step by step, and finally answer YES or NO.\\
   \hline
    \end{tabular}
    \vspace{-0.5em}
    \caption{Prompt for checking redundancy of merged candidate EGs.} \label{tab:redundancy_prompt}
\end{center}
\end{table}

\begin{table}[t]
\small
\begin{center}
\setlength{\tabcolsep}{5pt} 
    \begin{tabular}{p{0.9\linewidth} }
    \hline
    \textbf{Prompt} \\ \hline
    Given the following claim: ``\{\{claim\}\}'',
    and evidence sentences prepended with indices:\\
    \{\{evidence text\}\} \\
    \\
    Select the best minimal non-redundant group of evidence sentences that fully supports the claim. Only output sentence indices, separated by comma. \\
    \\
    Answer:\\
   \hline
    \end{tabular}
    \vspace{-0.5em}
    \caption{Prompt for directly predicting MEG.} \label{tab:direct_prompt}
\end{center}
\end{table}

\begin{table}[t]
\small
\begin{center}
\setlength{\tabcolsep}{5pt} 
    \begin{tabular}{p{0.9\linewidth} }
    \hline
    \textbf{Prompt} \\ \hline
    Write a claim that is fully supported by the given following evidence sentences: \\
    \{\{evidence text\}\} \\
   \hline
    \end{tabular}
    \vspace{-0.5em}
    \caption{Prompt for claim reconstruction.} \label{tab:claim_reconstruction}
\end{center}
\end{table}

\section{Base Model Performance} \label{sec:base_model}
\paragraph{Experimental settings.} To assess the support prediction \textit{Model} performance, we construct datasets of 2255 and 462 entailment examples respectively from WiCE test-set and SciFact dev-set. The sampled WiCE subset contains 1139, 322, 794 \textit{fully support}, \textit{partially support}, and do \textit{not support} examples, respectively. We directly use the annotated EGs from \textit{fully} and \textit{partially supporting} examples as inputs and randomly sample 1$\sim$3 EPs to serve as negative labels in \textit{not supporting} examples. Similarly for SciFact, we treat each annotated evidence group as \textit{fully supporting} and the subsets of annotated evidence groups as \textit{partially supporting}; we randomly sample 1$\sim$3 non-annotated EPs to as negative lables for \textit{not supporting} examples, obtaining 216, 42, and 204 \textit{fully support}, \textit{partially support}, and do \textit{not support} examples, respectively. Table \ref{tab:base_model} shows the prompt used for the LLM.

\paragraph{Experimental results.} Table \ref{tab:base_model} shows the support prediction \textit{Model} performance. Overall the model yields satisfactory performance on \textit{fully} and \textit{not supporting} examples but performs poorly on \textit{partially supporting} examples. This is because the \textit{partial support} label is vaguely defined, and presumably the LLM \cite{anil2023palm} did not encounter sufficient partially supporting entailment examples in its pretraining.

\section{Implementation Details} 
\subsection{Additional Preprocessing}
For the WiCE-MEG dataset, since the majority of the candidate EPs are not relevant to the claim, but some may be selected as part of the EGs by the LLM, we additionally filter out sentences without any stemmed token overlap with the claim in advance. This filtering removes 55.6\% of candidate EPs but affects only 6.7\% of gold EGs, significantly speeding up inference with minimal performance loss.

\subsection{Detailed Algorithm} \label{sec:implementation}
To avoid redundant computation, we iteratively merge two \textit{partially supporting} set of EPs to a larger candidate set and store it in $PGs$ in Algorithm \ref{alg:evidence_group_identification}. Therefore, $PGs$ is implemented by a Python dictionary with size of the set of EPs as keys and another nested Python dictionary $CS$ as values. Each $CS$ has a key of the merged set of EPs $G_1 \cup G_2$, and values of pair of the $(G_1, G_2)$. Algorithm \ref{alg:evidence_group_identification} \& \ref{alg:merge} presents the full pseudo code of our implementation. In Algorithm \ref{alg:merge} we prepare non-redundant candidate sets of EPs by running $notRedundant$ checker implemented by a zero-shot LLM prompt (Table \ref{tab:redundancy_prompt}).

\subsection{Inter-annotator Disagreement}
In WiCE \cite{kamoi-etal-2023-wice} dataset, we observe some inter-annotator disagreements where some human-labeled EGs are supersets of the other EGs for the same claim, but in these cases we still include both EGs as references.

\begin{algorithm}[t] \small
\caption{Minimal Evidence Group Identification with a support prediction Model.}\label{alg:evidence_group_identification}
\begin{algorithmic}
\Require $c$, $E=\{e_1, e_2, ..., e_n\}$, $Model$
\Require $max\_size$ \Comment{Max size of EGs to consider.}
\State $MEG \gets []$  \Comment{Proposed MEGs.}
\State $PGs \gets \{\}$  \Comment{Dict[size: Dict[G: \{G\}]]}
\For{$size$ in $1 ... \min(|E|, max\_size)$}
\State $PGs \gets MergePartialGroup(c, E, size, PGs)$ 
\State $CS \gets PGs[size].keys()$ \Comment{All candidate sets of EPs with size $size$}
\For{$S$ in $CS$}
\State $label \gets Model(c, S)$ 
\If{$label$ is $fully\ support$}
    \State $MEG.append(S)$
    \State \textbf{pop} $PGs[size][S]$
\ElsIf{$label$ is $not\ support$}
    \State \textbf{pop} $PGs[size][S]$
     
\EndIf
\EndFor
\If {$len(MEG)>0$}
\textbf{break}
\EndIf
\EndFor
\State \textbf{Output}  $MEG$
\end{algorithmic}
\end{algorithm}

\begin{algorithm}[t] \small
\caption{Merging partial evidence groups with redundancy checking.}\label{alg:merge}
\begin{algorithmic}
\Require $notRedundant$ \Comment{Redundancy Checker.}
\Function{MergePartialGroup}{$c$, $E$, $size$, $PGs$}
\State $CS \gets \{\}$ \Comment{Dictionary of Sets.}

\If{$size$ = 1}
    \For{$e$ in $E$}
    \State $CS[(e,)] \gets set([])$ 
    \EndFor
\Else
    \ForEach{pair $G_1 \in PGs[|G_1|]$ \& $G_2 \in PGs[|G_2|]$ s.t. $|G_1 \cup G_2|=size$ \& $notRedundant(c, G_1, G_2)$ }
    \State $CS[G_1 \cup G_2].add((G_1, G_2))$
    \EndFor
\EndIf
\State $PGs[size] \gets CS$
\State \Return \textit{$PGs$}
\EndFunction
\end{algorithmic}
\end{algorithm}

\end{document}